%% file: main.tex
\pdfoutput=1
\documentclass[11pt]{article}
\usepackage{emnlp2021}
\usepackage{times}
\usepackage{latexsym}
\usepackage[T1]{fontenc}

\usepackage[utf8]{inputenc}
\usepackage{enumitem}
\usepackage{graphicx}
\usepackage{bbding}
\usepackage{microtype}
\usepackage{url}
\usepackage{amsmath}
\usepackage{amsfonts}
\usepackage{multirow}
\usepackage{booktabs}
\usepackage{caption}
\usepackage{subfigure}
\usepackage{appendix}
\usepackage{xspace}
\newcommand{\tabref}[1]{Table~\ref{#1}\xspace}
\newcommand{\appref}[1]{Appendix\xspace\ref{#1}\xspace}

\newcommand{\sentence}{\mathbf{s}}
\newcommand{\word}{\mathbf{e}}
\newcommand{\aspect}{\mathbf{a}}
\newcommand{\wordsentimask}{\mathbf{p}_w}
\newcommand{\sensentimask}{\mathbf{p}_s}
\newcommand{\wordhidden}{h}

\newcommand{\senembed}{\mathbf{f}}
\newcommand{\sentilabel}{\mathbf{y}}

\newcommand{\generator}{\mathbf{G}}
\newcommand{\discriminator}{\mathbf{D}}
\newcommand{\ourmodel}{\text{SentiWSP}}

\title{Sentiment-Aware Word and Sentence Level Pre-training for Sentiment Analysis}

\author{
Shuai Fan\textsuperscript{\rm 1},\ \ 
Chen Lin\textsuperscript{\rm 1}\footnotemark[1],\ \ 
Haonan Li\textsuperscript{\rm 2}\footnotemark[2],\ \
Zhenghao Lin\textsuperscript{\rm 1}\footnotemark[2],\ \
Jinsong Su\textsuperscript{\rm 1}
Hang Zhang\textsuperscript{\rm 3},\\
\textbf{
Yeyun Gong\textsuperscript{\rm 4},
Jian Guo\textsuperscript{\rm 3},
Nan Duan\textsuperscript{\rm 4}}
\\
\textsuperscript{\rm 1} 
School of Informatics, Xiamen University, China\\
\textsuperscript{\rm 2}
The University of Melbourne, Australia\\
\textsuperscript{\rm 3}
IDEA Research, China\\
\textsuperscript{\rm 4}
Microsoft Research Asia
}

\begin{document}
\maketitle
\renewcommand{\thefootnote}{\fnsymbol{footnote}}
 \footnotetext[1]{Corresponding author, chenlin@xmu.edu.cn}
 \footnotetext[2]{Equal contribution}
\begin{abstract}
Most existing pre-trained language representation models (PLMs) are sub-optimal in sentiment analysis tasks, as they capture the sentiment information from word-level while under-considering sentence-level information. 
In this paper, we propose $\ourmodel$, a novel \textbf{Senti}ment-aware pre-trained language model with combined \textbf{W}ord-level and \textbf{S}entence-level \textbf{P}re-training tasks.
The word level pre-training task detects replaced sentiment words, via a generator-discriminator framework, to enhance the PLM's knowledge about sentiment words.
The sentence level pre-training task further strengthens the discriminator via a contrastive learning framework, with similar sentences as negative samples, to encode sentiments in a sentence.
Extensive experimental results show that $\ourmodel$ achieves new state-of-the-art performance on various sentence-level and aspect-level sentiment classification benchmarks.  
We have made our code and model publicly available at https://github.com/XMUDM/SentiWSP.

\end{abstract}

\input{introduction}
\input{related_work}
\input{model}

\input{experiment}
\input{conclusion}

\bibliography{main}
\bibliographystyle{acl_natbib}

\clearpage
\input{appendix}
\end{document}

%% file: introduction.tex
\section{Introduction}
Sentiment analysis plays a fundamental role in natural language processing (NLP) and powers a broad spectrum of important business applications such as marketing~\cite{sentimarket19} and campaign monitoring~\cite{Sandoval-Almazan20a}.
Two typical sentiment analysis tasks are sentence-level sentiment classification~\cite{bertpt19,sentibert20,TangXWHGCZX22} and aspect-level sentiment classification~\cite{SCAPT21}.

Recently, pre-trained language representation models (PLMs) such as ELMo~\cite{elmo2018}, GPT~\cite{gpt2018,gpt2019}, BERT~\cite{bert2019}, RoBERTa~\cite{robert19} and XLNet~\cite{xlnet19} have brought impressive performance improvements in many NLP problems, including sentiment analysis.  
PLMs learn a robust encoder on a large unlabeled corpus, through carefully designed pre-training tasks, such as masked token or next sentence prediction.

Despite their progress, the application of general purposed PLMs in sentiment analysis is limited, because they fail to distinguish the importance of different words to a specific task. 
For example, it is shown in~\cite{negated20} that general purposed PLMs have difficulties dealing with contradictory sentiment words or negation expressions, which are critical in sentiment analysis. 
To address this problem, recent sentiment-aware PLMs introduce  
word-level sentiment information, such as token sentiments and emoticons~\cite{sentix20}, aspect word~\cite{SKEP20}, word-level linguistic knowledge~\cite{sentilare20}, and implicit sentiment-knowledge information~\cite{SCAPT21}.
These word-level pre-training tasks, e.g., sentiment word prediction and word polarity prediction, mainly learn from the masked words and are not efficient to capture word-level information for all input words. Furthermore, sentiment expressed in a sentence is beyond the simple aggregation of word-level sentiments. However, general purposed PLMs and existing sentiment-aware PLMs underconsider sentence-level sentiment information.

In this paper, we propose a novel sentiment-aware pre-trained language model called $\ourmodel$, to combine word-level pre-training and sentence-level pre-training. 
Inspired by ELECTRA~\cite{electra20}, which pre-trains a masked language model with significantly less computation resource, we adopt a generator-discriminator framework in the word-level pre-training. The generator aims to replace masked words with plausible alternatives; and the discriminator aims to predict whether each word in the sentence is an original word or a substitution. To tailor this framework for sentiment analysis, we mask two types of words for generation, sentiment words and non-sentiment words. We increase the portion of masked sentiment words so that the model focuses more on the sentiment expressions.

For sentence-level pre-training, we design a contrastive learning framework to improve the encoded embeddings by the discriminator. 
The query for the contrastive learning is constructed by masking sentiment expressions in a sentence. The positive example is the original sentence. The negative examples are selected firstly from in-batch samples and then from cross-batch similar samples using an asynchronously updated approximate nearest neighboring (ANN) index. 
In this way, the discriminator, which will be used as the encoder for downstream tasks,
learns to distinguish different sentiment polarities even if they are superficially similar.

Our main contributions are in three folds: 1). $\ourmodel$ strengthens word-level pre-training via masked sentiment word generation and detection, which is more sample-efficient and benefits various sentiment classification tasks; 2). $\ourmodel$ combines word-level pretraining with sentence-level pre-training, which has been underconsidered in previous studies. SentiWSP adopts contrastive learning in the pre-training, where sentences are progressively contrasted with in-batch and cross-batch hard negatives, so that the model is empowered to encode detailed sentiment information of a sentence; 3). We conduct extensive experiments on sentence-level and aspect-level sentiment classification tasks, and show that $\ourmodel$ achieves new state-of-the-art performance on multiple benchmarking datasets.
\begin{figure*}[!htbp]
	\centering
	\includegraphics[width=1\textwidth]{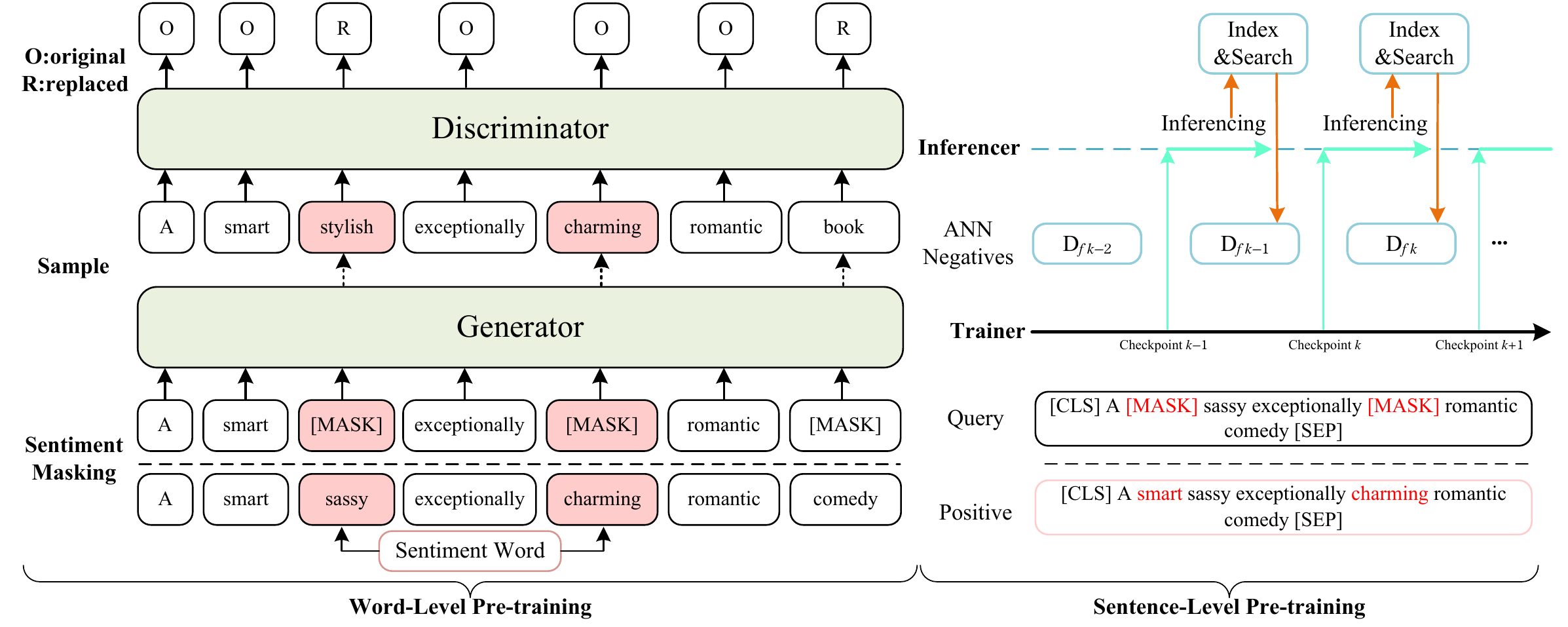}
	\caption{Framework overview of SentiWSP.}
	\label{fig:framwork}
\end{figure*}

%% file: related_work.tex
\section{Related Work}

\paragraph{Pre-training and Representation Learning}
Pre-training models has shown great success across various NLP tasks~\cite{bert2019, xlnet19, robert19}. 
Existing studies mostly use a Transformer-based~\cite{vaswani2017attention} encoder to capture contextual features, along with masked language model (MLM) and/or next sentence prediction~\citep{bert2019} as the pre-training tasks. 
\citet{xlnet19} propose XLNet which is pre-trained using a generalized autoregressive method that enables learning bidirectional contexts by maximizing the expected likelihood over all permutations of the factorization order.
ELECTRA~\cite{electra20} is a generator-discriminator framework, where the generator performs the masked token generation and the discriminator performs replaced token detection pre-training task. It is more efficient than MLM because the discriminator models over all input tokens rather than the masked tokens only. 
Our work improves ELECTRA's performance on sentiment analysis tasks, by specifying masked sentiment words at word-level pre-training, and combining sentence-level pre-training.

In addition to the pre-training models that encode token representations, sentence-level and passage-level representation learning have undergone rapid development in recent years.
A surge of work demonstrates that contrastive learning is an effective framework for sentence- and passage-level representation learning~\cite{COCO-LM,InfoXLM,SimCSE,KFCNet}.
The common idea of contrastive learning is to pull together an anchor and a ``positive'' sample in the embedding space, and push apart the anchor from ``negative'' samples.
Recently, COCO-LM~\cite{COCO-LM} creates positive samples by masking and cropping tokens from sentences. 
\citet{SimCSE} demonstrate that constructing positive pairs with only standard dropout as minimal data augmentation works surprisingly well on the Natural Language Inference (NLI) task.
\citet{dpr20} investigate the impact of different negative sampling strategies for passage representation learning based on the task of passage retrieval and question answering.
ANCE~\cite{ance21} adopts approximate nearest neighbor negative contrastive learning, a learning mechanism that selects hard negatives globally from the entire corpus, using an asynchronously updated Approximate Nearest Neighbor (ANN) index. 
Inspired by COCO-LM~\cite{COCO-LM} and ANCE~\cite{ance21}, we construct positive samples by masking a span of words from a sentence, and construct cross-batch hard negative samples to enhance the discriminator at sentence-level pre-training.

\paragraph{Pre-trained Models for Sentiment Analysis}
In the field of sentiment analysis, BERT-PT~\cite{bertpt19} conducts post-training on the corpora which belong to the same domain of the downstream tasks to benefit aspect-level sentiment classification. 
SKEP~\cite{SKEP20} constructs three sentiment knowledge prediction objectives in order to learn a unified 
sentiment representation for multiple sentiment analysis tasks. 
SENTIX~\cite{sentix20} investigates domain-invariant sentiment knowledge from large-scale review datasets, and utilizes it for cross-domain sentiment classification tasks without fine-tuning.  
SentiBERT~\cite{sentibert20} proposes a two-level attention mechanism on top of the BERT representation to capture phrase-level compositional semantics. 
SentiLARE~\cite{sentilare20} devises a new pre-training task called label-aware masked language model to construct knowledge-aware language representation. 
SCAPT~\cite{SCAPT21} captures both implicit and explicit sentiment orientation from reviews by aligning the representation of implicit sentiment expressions to those with the same sentiment label.

%% file: model.tex
\section{Sentiment-Aware Word-Level and Sentence-Level Pre-training }

The overall framework of SentiWSP is depicted in Figure~\ref{fig:framwork}. SentiWSP consists of two pre-training phases, namely Word-level pre-training (Sec.~\ref{sec:wordtask}, and Sentence-level pre-training (Sec.~\ref{sec:sentencetask}), before fine-tuning (Sec.~\ref{sec:finetune}) on a downstream sentiment analysis task.

In \textbf{word-level pre-training}, an input sentence flows through a word-masking step, followed by a generator to replace the masked words, and a discriminator to detect the replacements. The generator and discriminator are jointly trained in this stage. 
Then, the training of discriminator continues in \textbf{sentence-level pre-training}. Each input sentence is masked at sentiment words to construct a query, while the original sentence is treated as the positive sample. Their embeddings encoded by the discriminator are contrasted to two types of negative samples constructed in an in-batch warm-up training step and a cross-batch approximate nearest neighbor training step. Finally, the discriminator is \textbf{fine-tuned} on the downstream task.

Compared with previous studies, the discriminator in SentiWSP has three advantages. (1) Instead of random token replacement and detection, SentiWSP masks a large portion of sentiment words, and thus the discriminator pays more attention to word-level sentiments. (2) Instead of pure masked token prediction, SentiWSP incorporates context information from all input words via a replacement detection task. (3) SentiWSP combines sentence-level sentiments with word-level sentiments by progressively contrasting a sentence with missing sentiments to a superficially similar sentence.

\subsection{Word-Level Pre-training} \label{sec:wordtask}
\paragraph{Word masking}

Different from previous random word masking \cite{bert2019,electra20}, our goal is to corrupt the sentiment of the input sentence. 

In detail, we first randomly mask $15\%$ words, the same as ELECTRA~\cite{electra20}. Then, we use SentiWordNet~\cite{sentiwordnet10} to mark the positions of sentiment words in a sentence, and mask the sentiment words until a certain proportion $\wordsentimask$ of sentiment words are hidden. We empirically find that the sentiment word masking proportion $\wordsentimask=50\%$ achieves the best results. 

In the example in Figure~\ref{fig:framwork} (left), the sentiment words ``sassy'' and ``charming'' are masked while ``smart'' is not masked, ``comedy'' is masked as a random non-sentiment word.

\paragraph{Generator}

Next, a generator $\generator$ processes the masked sentence and generates a corrupted sentence. As in ELECTRA~\cite{electra20}, the generator is a small sized Transformer. Formally, the sentence is a sequence of words, i.e., $ \sentence= [w_1, w_2,\ldots, w_n]$, the mask indicators are denoted as  $\mathbf{m}=[m_1,m_2,\ldots,m_n],m_t\in\{0,1\}$, we obtain $\sentence^{\text{mask}}$ from the word masking step. For masked out words, the word is replaced by ``MASK", i.e., $\forall m_t=1,w_t=[``MASK"]$. 
The generator $\generator$ encodes the input to contextualized representations $\wordhidden_{\generator}(\sentence^{\text{mask}})=[\wordhidden_1,\ldots,\wordhidden_n]$.

For a given position $t$, (in our case only positions where $w_t=[``MASK"]$), the generator $\generator$ outputs a probability $p_{\generator}\left(w_t \mid \sentence^{\text{mask}} \right)$ for generating a particular token $w_t$ with a softmax layer:
\begin{equation}\label{equ:genword}
p_{\generator}\left(w_t \!\mid\! \sentence^{\text{mask}} \right)\!=\!\frac{\exp \left(\word_t^{T}\! \wordhidden_{\generator}(\sentence^{\text{mask}})_{t}\right)} {\sum_{w_j}\!\exp \left(\word_j^{T} \wordhidden_{\generator}(\sentence^{\text{mask}})_{t}\right)}
\end{equation}

where $\word_t$ denotes word embeddings for word $w_t$. 

We then replace the current word $w_t$ with a random sample strategy based on $p_{\generator}\left(w_t \mid \sentence^{\text{mask}} \right)$. Sampling introduces randomness and thus it is beneficial for training the discriminator. On the contrary, selecting the word with the highest probability is likely to generate the original word, and the training for discriminator will be more challenging as the discriminator is likely to be trapped to distinguish an original word from a substitution. Formally, the replacing process can be defined as $\forall m_t=1,\hat{w}_{t} \sim p_{\generator}\left(w_{t} \mid \sentence^{\text{mask}}\right)$. We denote the corrupted sequence as $\sentence^{\text{rep}}=[{w}_{1}^{\text{rep}},\ldots,{w}_{n}^{\text{rep}}]$, where $\forall m_t=1,w_t^{\text{rep}}=\hat{w}_t$.

\paragraph{Discriminator}
For the corrupted sentence, the discriminator $\discriminator$, i.e., a larger sized Transformer, encodes the corrupted sentence to $\wordhidden_{\discriminator}(\sentence^{\text{rep}})$, and predicts whether each word $w_t$ comes from the data or the generator, using a sigmoid output layer:
\begin{equation}\label{equ:discword}
\discriminator(\sentence^{\text{rep}}, t)=\sigma\left(\word_t^{T} \wordhidden_{\discriminator}(\sentence^{\text{rep}})_{t}\right)
\end{equation}
We jointly train the generator and the discriminator. The generator $\generator$ is trained by maximal likelihood estimation, and the  discriminator $\discriminator$ is trained by cross entropy.
\begin{align}
    \min _{\theta_{\generator}, \theta_{\discriminator}} & \sum_{\sentence \in \mathcal{X}} \mathcal{L}_{\generator}\left(\sentence, \theta_{\generator}\right)+\lambda \mathcal{L}_{\discriminator}\left(\sentence, \theta_{\discriminator}\right) \\\nonumber
    \mathcal{L}_{\generator}&=\sum_{m_t=1}-\log p_{\generator}\left(w_{t} \mid \sentence^{\text {mask}}\right) \label{equ:genloss} \\\nonumber
    \mathcal{L}_{\discriminator}&=\sum_{i=1}^{n}- \mathbb{I}(w_i^{\text{rep}}=w_i)\log \discriminator(\sentence^{\text{rep}}, i) \\\nonumber
    &- \big(1- \mathbb{I}(w_i^{\text{rep}}=w_i)\big)\log \left(1-\discriminator(\sentence^{\text{rep}}, i)\right) 
\end{align}\label{equ:allloss}
where $\mathcal{X}$ denotes a large corpus of raw text and $\lambda$ is the coefficient of the discriminator loss.  

\begin{table*}[htbp]
\resizebox{\textwidth}{!}{
	\centering
	\begin{tabular}{l|ccccc|cccc}
		\toprule
		\multirow{3}{*}{\centering{\bf Model}}       & \multicolumn{5}{c|}{\textbf{Sentence-level}}                                      & \multicolumn{4}{c}{\textbf{Aspect-level}}                                      \\ 
		& \textbf{IMDB}  & \textbf{SST-5} & \textbf{Yelp-2} & \textbf{Yelp-5} & \textbf{MR}    & \multicolumn{2}{c}{\textbf{Restaurant14}} & \multicolumn{2}{c}{\textbf{Laptop14}} \\
		& \textbf{Acc}   & \textbf{Acc}   & \textbf{Acc}    & \textbf{Acc}    & \textbf{Acc}   & \textbf{Acc}        & \textbf{MF1}        & \textbf{Acc}      & \textbf{MF1}      \\ 
		\midrule
		BERT~\cite{bert2019}              & 93.87            & 53.37            & 97.74            & 70.16            & 87.52            & 83.77              & 76.06                & 78.53            & 73.11              \\
		XLNet~\cite{xlnet19}             & 96.21            & 56.33            & 97.41            & 70.23            & 89.45            & 84.93              & 76.70                & 80.00            & 75.88              \\
		RoBERTa~\cite{robert19}           & 94.68            & 54.89            & 97.98            & 70.12            & 89.41            & 86.07              & 79.21                & 81.03            & 77.16              \\
		BERT-PT~\cite{bertpt19}           & 93.99            & 53.24            & 97.77            & 69.90            & 87.30            & 85.86              & 77.99                & 78.46            & 73.82              \\
		TransBERT~\cite{transbert19}         & 94.79            & 55.56            & 96.73            & 69.53            & 88.69            & 86.38              & 78.95                & 80.06            & 75.43              \\
		SentiBERT~\cite{sentibert20}         & 94.04            & 56.87            & 97.66            & 69.94            & 88.59            & 83.71              & 75.42                & 76.87            & 71.74              \\
		SentiLARE~\cite{sentilare20}         & 95.71            & 58.59            & 98.22            & 71.57            & 90.82            & 88.32              & 81.63                & 82.16            & 78.70              \\
		SENTIX~\cite{sentix20}            & 94.78            & 55.57            & 97.83            & \textbackslash{} & \textbackslash{} & 87.32              & \textbackslash{}     & 80.56            & \textbackslash{}   \\
		SCAPT~\cite{SCAPT21}             & \textbackslash{} & \textbackslash{} & \textbackslash{} & \textbackslash{} & \textbackslash{} & \textbf{89.11}     & \textbf{83.79}       & 82.76            & 79.15              \\
		ELECTRA~\cite{electra20}     & 95.62            & 57.89            & 97.87            & 71.27            & 90.81            & 87.32              & 81.63                & 82.13            & 78.25              \\
		\textbf{\ourmodel} & \textbf{96.26}   & \textbf{59.32}   & \textbf{98.25}   & \textbf{71.69}   & \textbf{92.41}   & 88.75              & 82.85                & \textbf{83.69}   & \textbf{80.82}     \\ 
		\bottomrule
	\end{tabular}}
    \caption{Overall performance of different models on sentiment classification tasks, ``Acc'' and ``MF1'' denote accuracy and macro-F1 score, respectively.}
	\label{tab:commpareresult}
\end{table*}

\subsection{Sentence-Level Pre-training}
\label{sec:sentencetask}
For sentence-level pre-training, we follow the contrastive framework in~\citet{SimCLR}. The goal of contrastive learning is to learn effective representations by pulling together similar samples (i.e., the positive samples) and pushing away different samples (i.e., the negative samples). 

One critical question  in contrastive learning is how to construct a pair of query (anchor) and positive/negative samples, i.e., $\left(q_{i}, d_{i}^{+}\right),\left(q_{i}, d_{i}^{-}\right)$. As the example shown in Figure~\ref{fig:framwork} (right), given a sentence $\sentence_i$ from corpus $C$, we first mask out a certain percentage (70\% in this research) of sentiment words in the sentence to construct $q_i$, and treat the raw sentence as the positive example $d_{i}^{+}$. 

\paragraph{In-batch warm-up training}

Then we fetch the already trained (in word-level pre-training) discriminator model $\discriminator$ and conduct a warm-up sentence-level training with in-batch negatives. 
In detail, We feed the input $\left(q_{i}, d_{i}^{+}\right)$ to the encoder $\discriminator$ to get the representations $\senembed_{i}$ and $\senembed_{i}^{+}$, and train the encoder to minimize the distance between the positive pairs within a mini-batch using the neg log-likelyhood loss defined as:
\begin{equation}\label{equ:warmup}
\min_{\theta_{\discriminator}}-\sum_{i\in\mathcal{B}}\log \frac{\exp( {\operatorname{sim}\left(\senembed_{i}, \senembed_{i}^{+}\right) / \tau})}{\sum_{j=1}^{|\mathcal{B}|} \exp( {\operatorname{sim}\left(\senembed_{i}, \senembed_{j}^{+}\right) / \tau})}
\end{equation}
where $\tau$ is a temperature hyperparameter, $|\mathcal{B}|$ denotes size of the mini-batch $\mathcal{B}$, $\operatorname{sim}\left(\cdot, \cdot\right)$ denotes cosine similarity between two vectors.

\paragraph{Cross-batch approximate nearest neighbor training}

Since in-batch negatives are unlikely to provide informative samples, we use the Approximate nearest neighbor Negative Contrastive Estimation, (ANCE)~\cite{ance21} to select ``hard'' negative samples from the entire corpus, to improve the discriminator's distinguishing power, using an asynchronously updated Approximate Nearest Neighbor (ANN) index.
In detail, after the warm-up training of the model $\discriminator$, we use it to infer the sentence embedding on the entire corpus $C$ and then use ANN search to retrieve top-$k$ negative examples closest to each query. 
Then we sample $t$ negative examples as hard-negative examples from the top-$k$ negatives. 
The hyperparameters $k$ and $t$ are set to 100 and 7, respectively.

To maintain an up-to-date ANN index two operations are required: (1) inference: refresh the embeddings of all sentences in the corpus with the updated model $\discriminator$; and (2) indexing: rebuild the ANN index with the updated embeddings. 
Although indexing is efficient~\cite{JohnsonDJ21}, inferential computing for each batch is expensive as it needs to be passed forward across the entire corpus.
In order to balance the time cost between inference and indexing, we use an asynchronous refresh mechanism similar to ~\citet{yibu20} and update the ANN index  every $m$ steps. 
As illustrated in the top right part in Figure~\ref{fig:framwork}, we construct a trainer to optimize $\discriminator$, and an inferencer that uses the latest checkpoint (e.g., checkpoint $k-1$) to recalculate the encoding $\senembed^{k-1}$ of the entire corpus and and update $\text{ANN}_{\senembed^{k-1}}$. 
Then, the trainer optimizes a cross-entropy objective function with negative samples generated from $\text{ANN}_{\senembed^{k-1}}$ and the original positive example pair $\left(q_i, d_i^{+}\right)$.
\begin{align}\label{equ:cesentence}
    \min_{\theta_{\discriminator}}\sum_{i\in\mathcal{B}^{k}} &-\log \big(\operatorname{sim}(\senembed_i,\senembed_i^{+}) \big) \\\nonumber
    &- \sum_{j \sim \text{ANN}_{\senembed^{k-1}}}log (1- \operatorname{sim}(\senembed_i,\senembed_j))
\end{align}
where $\mathcal{B}^{k}$ is the mini-batch batch at checkpoint $k$, $\senembed_i,\senembed_i^{+},\senembed_j$ indicate the discriminator $\discriminator$'s embeddings of the query, positive, and negative samples generated from the asynchronously updated ANN, respectively. 

\begin{table*}[htbp]
	\centering
	\begin{tabular}{l|ccccc|cccc}
		\toprule
		\multirow{3}{*}{\centering{\bf Model}}       & \multicolumn{5}{c|}{\textbf{Sentence-level}}                                      & \multicolumn{4}{c}{\textbf{Aspect-level}}                                      \\ 
		& \textbf{IMDB}  & \textbf{SST-5} & \textbf{Yelp-2} & \textbf{Yelp-5} & \textbf{MR}    & \multicolumn{2}{c}{\textbf{Restaurant14}} & \multicolumn{2}{c}{\textbf{Laptop14}} \\
		& \textbf{Acc}   & \textbf{Acc}   & \textbf{Acc}    & \textbf{Acc}    & \textbf{Acc}   & \textbf{Acc}        & \textbf{MF1}        & \textbf{Acc}      & \textbf{MF1}      \\ 
		\midrule
		\textbf{\ourmodel-base}  & \textbf{95.57} & \textbf{58.12} & \textbf{98.08}  & \textbf{71.09}  & \textbf{90.46} & \textbf{87.14}      & \textbf{81.33}      & \textbf{82.12}    & \textbf{78.34}    \\ 
		\quad w/o WP          & 95.46          & 57.64          & 98.02           & 70.82           & 90.17          & 86.82               & 80.02               & 81.83             & 77.87             \\
		\quad w/o SP         & 95.41          & 57.97          & 98.01           & 70.78           & 90.12          & 86.75               & 79.98               & 81.75             & 77.54             \\
		\quad w/o WP,SP          & 94.98          & 57.17          & 97.59           & 70.67           & 89.82          & 86.13               & 79.53               & 81.35             & 77.23             \\
		\midrule
		\textbf{\ourmodel-large} & \textbf{96.26} & \textbf{59.32} & \textbf{98.25}  & \textbf{71.69}  & \textbf{92.41} & \textbf{88.75}      & \textbf{82.85}      & \textbf{83.69}    & \textbf{80.82}    \\ 
		\quad w/o WP       & 96.12          & 58.67          & 98.21           & 71.31           & 91.87          & 87.87               & 82.13               & 82.97             & 79.35             \\
		\quad w/o SP        & 96.17          & 58.34          & 98.17           & 71.35           & 91.77          & 87.56               & 81.85               & 82.89             & 79.13             \\
		\quad w/o WP,SP   & 95.62          & 57.89          & 97.87           & 71.27           & 90.81          & 87.32               & 81.63               & 82.13             & 78.25             \\
		\bottomrule
	\end{tabular}
	\caption{The ablation study results. SP and WP represent sentence-level pre-training and word-level pre-training, respectively. The model without both pre-training is the original ELECTRA model.}
	\label{tab:ablationresult}
\end{table*}

\subsection{Fine-tuning}\label{sec:finetune}
After the pre-training, we fine-tune our model on downstream sentiment analysis tasks.
For sentence-level sentiment classification task, we format the input sequence as:
$[CLS],\word_1,\ldots,\word_n,[SEP]$, and take the representation at the $[CLS]$ token to predict the sentiment label $\sentilabel$, which indicates the sentiment polarity of the sentence. 

For aspect-level sentiment classification task, we format the input sequence as:
$[CLS],\aspect_1,\ldots,\aspect_m,[SEP],\word_1,\ldots,\word_n,[SEP]$. 
where $\aspect_1,\ldots,\aspect_m$ denotes the phrase of a particular aspect. We fetch the representation at the $[CLS]$ token to predict the sentiment label $\sentilabel$ of the sentence in the aspect.

%% file: experiment.tex
\begin{table}[t]
	\small
	\centering
	\begin{tabular}{llcc}
		\toprule
		\textbf{Dataset} & \textbf{Train/Valid/Test} & \textbf{Length} & \textbf{\#C} \\ 
		\midrule
		SST              & 8,544 / 1,101 / 2,210     & 19.2            & 5           \\
		MR               & 8,534 / 1,078 / 1,050     & 21.7            & 2           \\
		IMDB             & 22,500 / 2,500 / 25,000   & 279.2           & 2           \\
		Yelp-2           & 504,000 / 56,000 / 38,000 & 155.3           & 2           \\
		Yelp-5           & 594,000 / 56,000 / 50,000 & 156.6           & 5           \\
		Res14            & 2,163 / 150 / 638         & 25.6            & 3           \\
		Lap14            & 3,452 / 150 / 676         & 30.2            & 3           \\ 
		\bottomrule
	\end{tabular}
	\caption{Statistics of datasets used in our experiments. \#C indicates the number of target classes in each dataset.}
	\label{tab:dataset}
\end{table}
\section{Experiment}

\subsection{Datasets}
For $\ourmodel$ pre-training, we use the same English Wikipedia corpus as \citet{bert2019}. 
We select 2 million sentences with a maximum length of 128 for the word-level pre-training, and select 500,000 sentences which have 20\%-30\% proportion of sentiment words for the sentence-level pre-training. 

After pre-training, we fine-tune our model on sentence-level sentiment classification benchmarks including Stanford Sentiment Treebank (SST)~\cite{sst2013}, IMDB~\cite{imdb11}, Movie Review (MR)~\cite{mr05}, and Yelp-2/5~\cite{yelp15}. 
For aspect-level sentiment classification tasks, we choose SemEval2014 Task 4 in laptop (Laptop14) and restaurant (Restaurant14) domains~\cite{semeval14}. 
Table~\ref{tab:dataset} shows the statistics of these datasets, including the amount of training, validation, and test splits, the average length and the number of classes. 
Since there is no validation set in MR, IMDB, and yelp-2/5, we randomly select a subset from the training set for validation.

\subsection{Baselines}
We compare our model with both general purposed pre-trained models and sentiment-aware pre-trained models.
For general purposed pre-trained models, we choose BERT~\cite{bert2019}, XLNet~\cite{xlnet19}, and RoBERTa~\cite{robert19} as baselines. 
For sentiment-aware pre-trained models, we choose BERT-PT~\cite{bertpt19}, SentiBERT~\cite{sentibert20}, SentiLARE~\cite{sentilare20}, SENTIX~\cite{sentix20}, and SCAPT~\cite{SCAPT21}. 
We also implement TransBERT~\cite{transbert19}, in the same transfer manner as SentiLARE.\footnote{We choose review-level sentiment classification on Yelp Dataset Challenge 2019 as the transfer task of TransBERT and the sentiment classification downstream tasks as the target tasks of TransBERT.}

\subsection{Implementation Details}
During pre-training, we use the AdamW optimizer and linear learning rate scheduler, and we set the max sequence length to 128.
The learning rate is initialized with 2e-5 and 1e-5 for the base and large model, respectively.
For word-level pre-training, we use ELECTRA~\cite{electra20} initialize $\generator$ and $\discriminator$. We set the proportion of sentiment word mask to $\wordsentimask=0.5$ and we keep other hyper-parameters the same as ELECTRA.
For sentence-level pre-training, we follow the settings of unsupervised SimCSE~\cite{SimCSE} to do the warm-up training, and set the proportion of sentiment word mask $\sensentimask=0.7$. 
The detailed batch size and training steps for different level of pre-training are listed in \appref{sec:parmdetail}.

For fine-tuning, we use the hyperparameters from~\citet{ClarkLKML19} for the most parts. 
We fine-tune 3-5 epochs for sentence-level sentiment classification and 7-10 epochs for aspect-level sentiment classification tasks. 
The learning rate of the base and large model for the fine-tunning is set to 2e-5 and 1e-5, respectively. 
We use a linear learning rate scheduler with 10\% warm-up steps.
\subsection{Comparative Results}
We list the performance of different models in Table~\ref{tab:commpareresult}. 
According to the results, we have several findings: 
(1) $\ourmodel$ consistently outperforms all baselines on sentence-level classification tasks, which demonstrates the superiority of $\ourmodel$ to capture sentence-level semantics. 
(2) On the aspect-level sentiment classification tasks, the proposed $\ourmodel$ boosts the ACC by $0.93$ and increases MF1 by $1.67$ on Laptop14 dataset. It also achieves a competitive performance on Resturant14 dataset, i.e., the second best among all competitors. 
(3) $\ourmodel$ is significantly better than ELECTRA, on both sentiment analysis tasks, on all datasets. This observation verifies the effectiveness of the proposed sentiment-aware pretraining strategy. 

\subsection{Ablation Study}
To further investigate the effectiveness of the combining word-level and sentence-level pre-training, we conduct an ablation study with different model sizes (details in Appendix~\ref{sec:parmdetail}). 
From the results (Table~\ref{tab:ablationresult}) we have the following observations:
(1) Without sentence-level pre-training and word-level pre-training (i.e., ``w/o WP,SP"), $\ourmodel$ degrades to ELECTRA. It performs worst in terms of all metrics. This proves the necessity of tailoring pre-training paradigms for sentiment analysis tasks.
(2) The full version of $\ourmodel$ produced the best results across different tasks and datasets. Word-level pre-training and sentence-level pre-training capture sentiment information at different granularity, and combining multi-granularity pre-training is beneficial.  
(3) Comparing sentence-level pre-training and word-level pre-training, $\ourmodel$ without sentence-level pre-training is generally worse than $\ourmodel$ without word-level pre-training. The performance decline is consistent on Aspect-level sentiment classification task. We think the reason is that the global context is essential for analyzing aspect sentiments, while focusing only on word-level information leads to less robust prediction.

\begin{table}[t]
	\centering
	\begin{tabular}{lc|cc}
		\toprule
		\textbf{Model}  & \textbf{$\wordsentimask$} & \textbf{IMDB}  & \textbf{MR}    \\ 
		\midrule
		\ourmodel-only WP & 0                  & 95.59          & 90.83          \\
		\ourmodel-only WP & 0.3                & 96.13          & 91.67          \\
		\ourmodel-only WP & 0.5                & \textbf{96.17} & \textbf{91.77} \\
		\ourmodel-only WP & 0.7                & 95.98          & 91.07          \\
		\ourmodel-only WP & 1                  & 95.97          & 91.26          \\ 
		\bottomrule
	\end{tabular}
	\caption{Acc obtained by additionally masking $\wordsentimask$ of sentiment words in word-level pre-training, on IMDB and MR.}
	\label{tab:wordparmresult}
\end{table}

\begin{table}[t]
	\centering
	\begin{tabular}{lc|cc}
		\toprule
		\textbf{Model}  & \textbf{$\sensentimask$} & \textbf{IMDB}  & \textbf{MR}    \\ 
		\midrule
		\ourmodel-only SP & 0.3                & 96.02          & 91.18          \\
		\ourmodel-only SP & 0.5                & 95.86          & 91.36          \\
		\ourmodel-only SP & 0.7                & \textbf{96.12} & \textbf{91.87} \\
		\ourmodel-only SP & 1                  & 95.89          & 91.61          \\ 
		\bottomrule
	\end{tabular}
	\caption{Acc obtained by additionally masking $\sensentimask$ of sentiment words in sentence-level pre-training, on IMDB and MR.}
	\label{tab:senparmresult}
\end{table}

\subsection{Impacts of Hyper-parameters}
In this section, we explore the impact of different hyper-parameters for the proposed model on IMDB and MR datasets.

\noindent\textbf{Word masking.}
For word-level pre-training, we experiment with replacing different proportions of sentiment words $\wordsentimask$ (Table~\ref{tab:wordparmresult}).
From the table, we find that the model performs the worst when $\wordsentimask = 0$ (i.e., the same mask strategy as ELECTRA)
, which verifies our assumption that extra sentiment word masking is beneficial for the model to encode sentiment information.
Besides, we find that masking and replacing $50\%$ of the sentiment words yields the best result. 
We argue the reason is that replacing $50\%$ sentiment words is difficult enough for the model to learn meaningful features, while keeping half of the original sentiment words provides useful clues for the model to detect other sentiment words.

\noindent\textbf{Sentence-level positive sample construction.}
Similar to the word-level experiment, we mask different proportions of sentiment words to construct positive samples for sentence-level pre-training. As shown in Table~\ref{tab:senparmresult}, the best performance is achieved by masking $70\%$ of the sentiment words. 
The underlying reason is that, the ideal positive sample should resemble the query, yet the augmentation can provide additional information for the model to learn meaningful representations. We believe $70\%$ of sentiment word masking is a good balancing point. It is worthy to point out that, even the worst performances, i.e., when $\sensentimask=0.5$ on IMDB and $\sensentimask=0.3$ on MR, are better than most of the competitors in Table~\ref{tab:commpareresult}.

\begin{table}[t]
	\centering
	\begin{tabular}{lc|cc}
		\toprule
		\textbf{Model} & $k$ & \textbf{IMDB}  & \textbf{MR}    \\ \midrule
		In-batch       & N/A                  & 95.85          & 91.60          \\
		ANCE           & 1                  & 95.97          & 91.69          \\
		ANCE           & 3                  & 96.15          & 91.87          \\
		ANCE           & 5                  & 96.12          & 91.79          \\
		ANCE           & 7                  & 96.26          & \textbf{92.41} \\
		ANCE           & 10                 & \textbf{96.27} & 92.12          \\
		ANCE           & 13                 & 96.19          & 91.83          \\\bottomrule
	\end{tabular}
	\caption{Acc obtained by using only in-batch negative samples and top-$k$ cross-batch negative samples.}
	\label{tab:negaparmresult}
\end{table}

\begin{table}[t]
	\centering
	\begin{tabular}{@{}cc|cc@{}}
        \toprule
        \textbf{Loss} & \textbf{Similarity} & \textbf{IMDB}  & \textbf{MR}    \\ \midrule
        Triplet       & Dot Product         & 96.05          & 91.95          \\
        Triplet       & Cosine              & 96.16          & 92.12          \\
        NLL           & Dot Product         & 96.21          & 92.27          \\
        NLL           & Cosine              & \textbf{96.26} & \textbf{92.41} \\ \bottomrule
    \end{tabular}
	\caption{The impact of loss and similarity functions in sentence-level pre-training.}
	\label{tab:lossablation}
\end{table}

\noindent\textbf{Negative sample size.}
In Table~\ref{tab:negaparmresult}, we report the results with different negative samples in the sentence-level pre-training. 
From the table we have two findings: (1) when only in-batch negative samples are used, the model performs worst. 
We argue the reason is that in-batch negatives are too simple for the model to distinguish  from a positive sample, and continue training on these ``easy'' negatives does not make further improvements. 
(2) When we increase the cross-batch negative sample size from $1$ to $10$, the  model is provided with more informative negative samples. Therefore, the model can learn more detailed sentiment information and the accuracy is improved.
However, when we use a large amount of cross-batch negative samples (e.g., $13$), the negative samples vary in quality, and thus the model suffers from less similar negatives. 

\noindent\textbf{Similarity and loss function.}
For sentence-level pre-training, we compare two commonly adopted similarity functions, i.e., cosine distance and dot product, to measure the similarity between two sentence embeddings.
The difference between dot product and cosine distance is that dot product does not incorporate L2 normalization.
We also compare two widely used loss functions, i.e., Negative Log-Likelihood (NLL) loss and Triplet loss \cite{Triplet15}, in contrastive learning and ranking problems. The difference between NLL loss and Triplet loss is that Triplet loss  compares a positive example and a negative one directly with respect to a query.
We report the comparisons in Table~\ref{tab:lossablation}.
From the table we have two observations: 
(1) With different loss functions, the cosine distance appears to be a more accurate measurement for sentence similarity and outperforms the dot product. 
(2) The NLL loss produces better results, with different similarity functions, than the Triplet loss.

\begin{figure}[t]
	\small
	\centering
	\includegraphics[width=0.8\columnwidth]{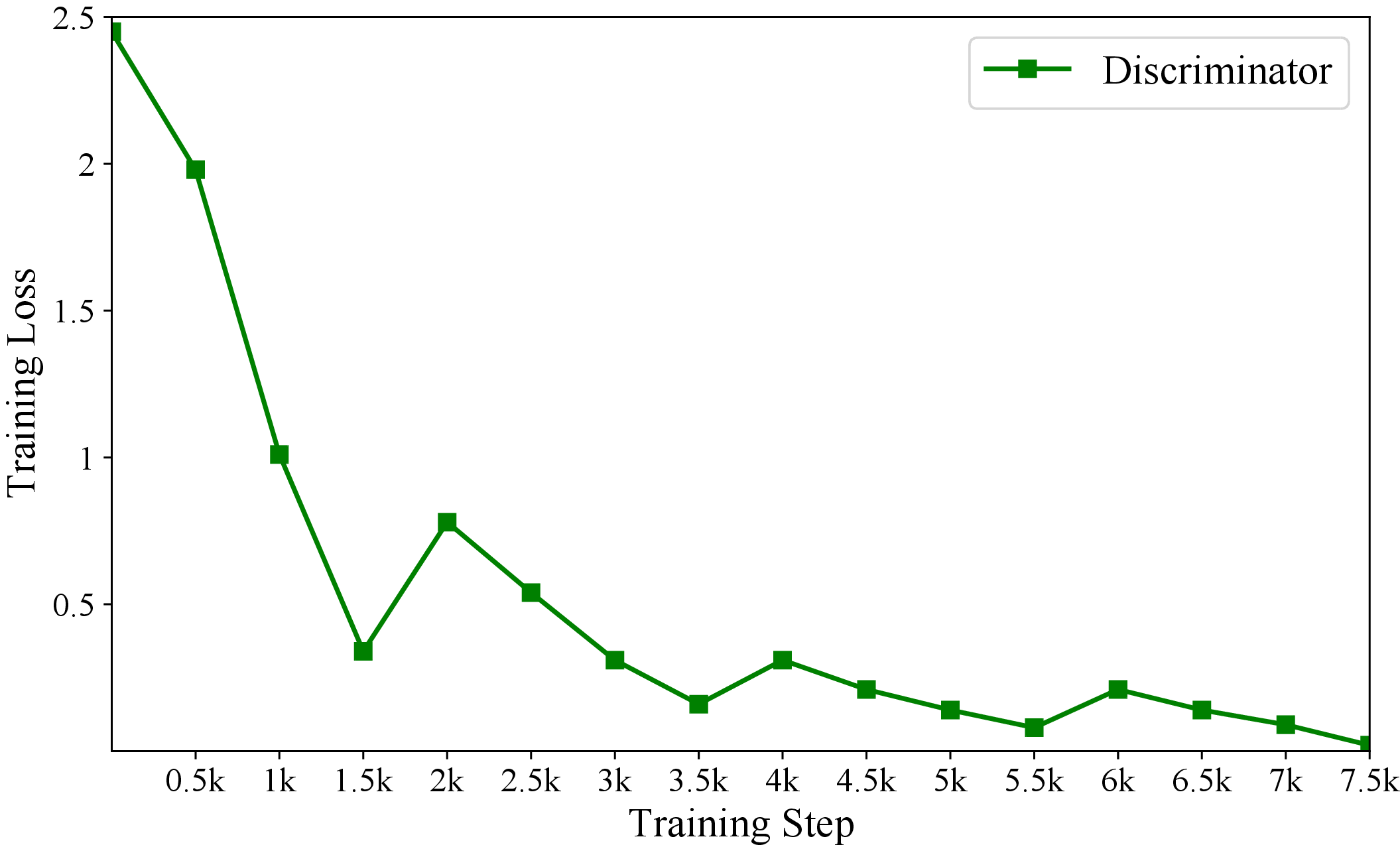}
	\caption{Training loss of sentence-level pre-training.}
	\label{fig:loss_sen}
\end{figure}

\begin{figure}[t]
	\centering
	\includegraphics[width=1\columnwidth]{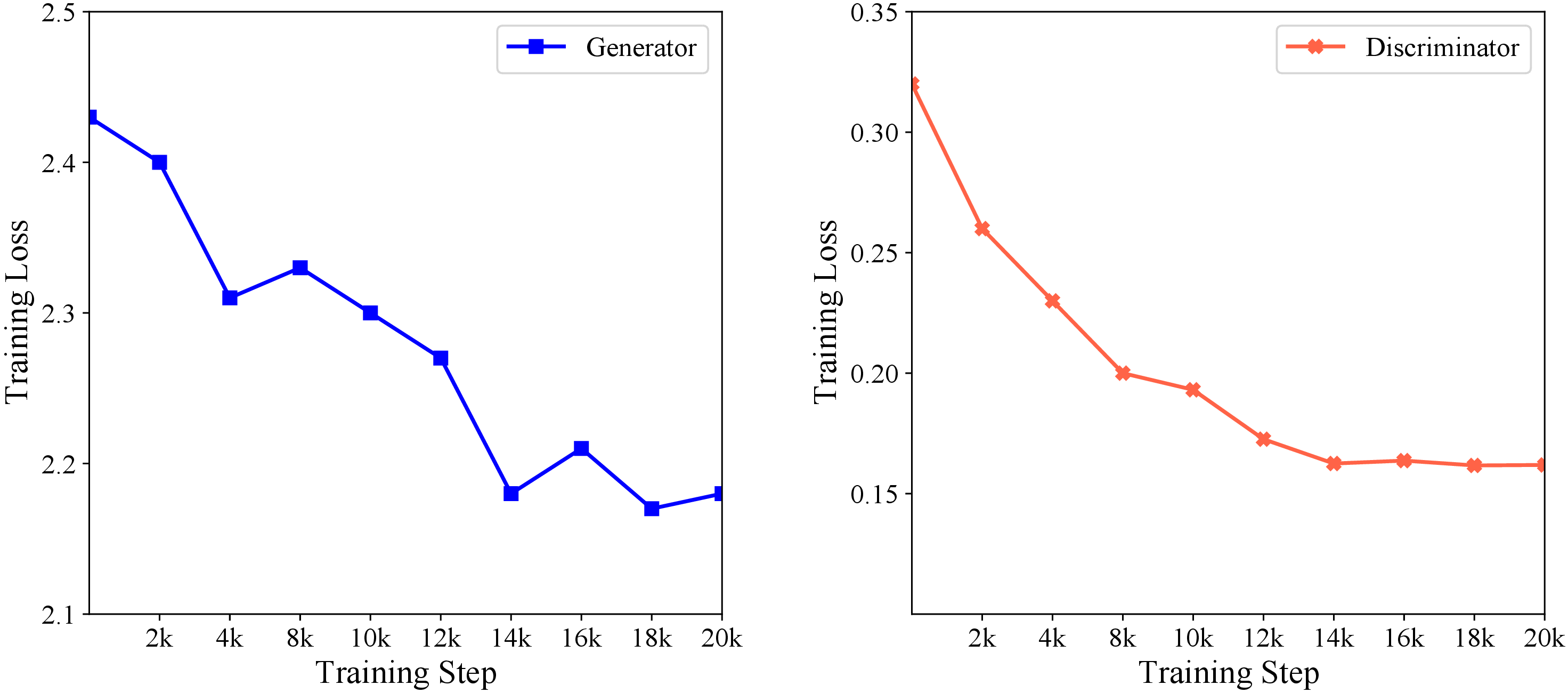}
	\caption{Training loss of the word-level pre-training.}
	\label{fig:loss_word}
\end{figure}

\subsection{Training Loss Convergence}
Our final model is trained on 4 NVIDIA Tesla A100 GPUs with a total training time of fewer than 24 hours.
For word-level pre-training, we can observe from Figure~\ref{fig:loss_word} that the generator and discriminator compete in joint training and gradually converge within $20,000$ steps.
For sentence-level pre-training, we can observe from Figure~\ref{fig:loss_sen} that when the hard-negative example is refreshed every 2000 steps, the loss of the model increases temporarily, which indicates that our ANN search can form a more demanding test for the model and improve the model's capability on these hard testing cases in the following steps.

%% file: conclusion.tex
\section{Conclusion}
In this paper, we introduce $\ourmodel$, which improves pre-training models on sentiment analysis task, by capturing the sentiment information from word-level and sentence-level simultaneously. 
Extensive experimental results on five sentence-level sentiment classification benchmarks show that $\ourmodel$ establishes new state-of-the-art performance on all of them. We conduct experiment on two aspect-level sentiment classification benchmarks. The results show that $\ourmodel$ beats most existing models on Restaurant14 and achieves new state-of-the-art on Laptop14.
We further analyze several hyper-parameters that may affect the model performance, and show that $\ourmodel$ can achieve satisfying performance with respect to different hyper-parameter settings.

\section*{Limitations}
$\ourmodel$, as most of the current state-of-the-art pre-training models, requires relatively large computation resources. As shown in Table~\ref{tab:ablationresult}, $\ourmodel$-large performs better than $\ourmodel$-base, the performance divergence is more significant on the MR dataset. We also observe some bad cases when sentiment expressions are implicit suggested in a sentence, i.e., with very few sentiment words, $\ourmodel$ has difficulty in masking and generating, and constructing positive samples. In the future, we plan to devise an adaptive masking mechanism for sentiment words.

\section*{Acknowledgements}
The project was supported by National Natural Science Foundation of China (No. 61972328, No. 62276219, No. 62036004), Natural Science Foundation of Fujian Province of China (No. 2020J06001), and Youth Innovation Fund of Xiamen (No. 3502Z20206059). We also thank the reviewers for their insightful comments. 

%% file: appendix.tex
\begin{appendices}

\section{Pre-train details}
\label{sec:parmdetail}
We provide detailed hyper-parameter settings in Table~\ref{tab:parmdetail} \tabref{tab:parmdetail}.
\begin{table}[h]
	\centering
    \begin{tabular}{l|c|cc}
        \hline
        \textbf{}& \textbf{Parameter} & \textbf{base} & \textbf{large} \\ \hline
        \multirow{3}{*}{\begin{tabular}[c]{@{}l@{}}Word-level\\ pretraining\end{tabular}}     
        & Batch size         & 128           & 64             \\
        & Warm up steps      & 1500          & 1500           \\
        & Max steps          & 20000         & 20000          \\ \hline
        \multirow{3}{*}{\begin{tabular}[c]{@{}l@{}}In-batch \\ Warm up\end{tabular}}          
        & Batch size         & 64            & 32             \\
        & Warm up steps      & 500           & 500           \\
        & Max steps          & 2000          & 2000           \\ \hline
        \multirow{4}{*}{\begin{tabular}[c]{@{}l@{}}Sentence-level\\ pretraining\end{tabular}} 
        & Batch size         & 64            & 32             \\
        & Iteration steps    & 2000          & 2000           \\
        & Max iterations     & 4             & 4              \\
        & Max steps          & 8000          & 8000           \\ \hline
    \end{tabular}
	\caption{SentiWSP pre-training.}
	\label{tab:parmdetail}
\end{table}
\end{appendices}